\documentclass[11pt,a4paper]{article}
\usepackage[hyperref]{conll2018}
\usepackage{times}
\usepackage{latexsym}
\usepackage{amsmath}
\usepackage{amssymb}
\usepackage{url}
\usepackage{multirow}
\usepackage{tikz}
\usepackage{tikz-qtree}

\usepackage{array}
\newcolumntype{P}[1]{>{\centering\arraybackslash}p{#1}}
\usepackage{graphicx}
\usepackage{caption}
\usepackage[export]{adjustbox}
\usepackage{lipsum}

\usepackage{url}

\aclfinalcopy % Uncomment this line for the final submission

\newcommand{\lform}[1]{\texttt{\fontsize{8.5pt}{.1pt}\selectfont{#1}}}

\title{Weakly-supervised Neural Semantic Parsing with a Generative Ranker}

\author{Jianpeng Cheng \and Mirella Lapata\\
Institute for Language, Cognition and Computation\\
School of Informatics, University of Edinburgh\\
10 Crichton Street, Edinburgh EH8 9AB\\
 {\tt
   \href{mailto:jianpeng.cheng@ed.ac.uk}{jianpeng.cheng@ed.ac.uk}}~~~~{\tt
   \href{mailto:mlap@inf.ed.ac.uk}{mlap@inf.ed.ac.uk}}
}
\date{}

\begin{document}
\maketitle
\begin{abstract}
  Weakly-supervised semantic parsers are trained on
  utterance-denotation pairs, treating logical forms as latent.  The
  task is challenging due to the large search space and spuriousness
  of logical forms. In this paper we introduce a neural parser-ranker
  system for weakly-supervised semantic parsing.
  The parser generates candidate tree-structured logical forms from utterances using clues of denotations.
  These candidates are then ranked based on two criterion: 
  their likelihood of executing to the correct denotation, and their
  agreement with the utterance semantics. 
  We present a scheduled training procedure to balance the contribution of the two objectives.
  Furthermore, we propose to use a neurally encoded lexicon to inject prior
  domain knowledge to the model.
  Experiments on three Freebase
  datasets demonstrate the effectiveness of our semantic parser,
  achieving results within the state-of-the-art range.
  
\end{abstract}

\section{Introduction}
\label{sec:introduction}

Semantic parsing is the task of converting natural language utterances
into machine-understandable meaning representations or logical forms.  The task has attracted much attention in the
literature due to a wide range of applications ranging from question
answering \cite{kwiatkowski2011lexical,liang2011learning} to relation
extraction \cite{krishnamurthy2012weakly}, goal-oriented dialog
\cite{wen-EtAl:2015:EMNLP}, and instruction understanding
\cite{chen2011learning,Matuszek:etal:2012,artzi-zettlemoyer:2013:TACL}.

In a typical semantic parsing scenario, a logical form
is executed against a knowledge base to
produce an outcome (e.g., an answer) known as denotation.
Conventional semantic parsers are trained on collections of utterances
paired with annotated logical forms
\cite{zelle1996learning,zettlemoyer:learning:2005,wong:learning:2006,kwiatkowksi2010inducing}. However,
the labeling of logical forms is labor-intensive and challenging to
elicit at a large scale.  As a result, alternative
forms of supervision have been proposed to alleviate the annotation
bottleneck faced by semantic parsing systems. One direction
is to train a semantic parser in a weakly-supervised setting
based on utterance-denotation pairs
\cite{clarke2010driving,kwiatkowski2013scaling,krishnamurthy2012weakly,cai2013large},
since such data are relatively easy to obtain via crowdsourcing
\cite{berant-EtAl:2013:EMNLP}.

\begin{figure}[t]
	\centering\includegraphics[width=3in]{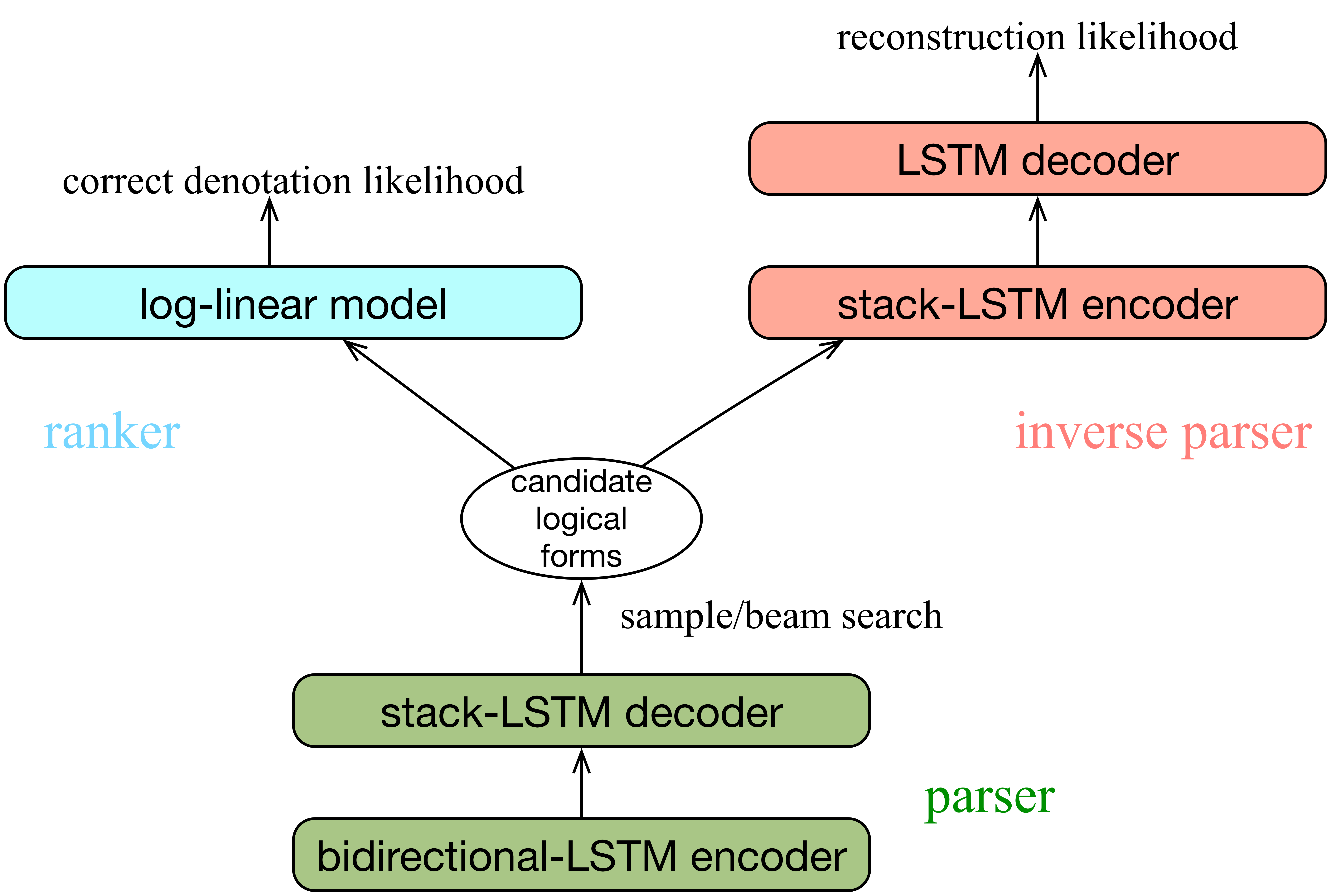}
	\caption{Overview of the weakly-supervised neural semantic parsing system.}\label{fig:1}
\end{figure}

However, the unavailability of logical forms in the weakly-supervised
setting, renders model training more difficult.  A fundamental
challenge in learning semantic parsers from denotations is finding
\emph{consistent} logical forms, i.e.,~those which execute to the
correct denotation. This search space can be very large, growing
exponentially as compositionality increases. Moreover, consistent
logical forms unavoidably introduce a certain degree of
\emph{spuriousness} --- some of them will accidentally execute to the
correct denotation without reflecting the meaning of the
utterance. These spurious logical forms are misleading
supervision signals for the semantic parser.

In this work we introduce a weakly-supervised neural semantic parsing
system which aims to handle both challenges.  Our system, shown in
Figure~\ref{fig:1}, mainly consists of a sequence-to-tree parser which
generates candidate logical forms for a given utterance.  These
logical forms are subsequently ranked by two components: a log-linear
model scores the likelihood of each logical form executing to the
correct denotation, and an inverse neural parser measures the degree
to which the logical form represents the meaning of the utterance.  We
present a scheduled training scheme which balances the contribution of
the two components and objectives.  To further boost performance, we
propose to neurally encode a lexicon, as a means of injecting prior
domain knowledge to the neural parameters.

We evaluate our system on three Freebase datasets which consist of
utterance denotation pairs: \textsc{WebQuestions}
\cite{berant-EtAl:2013:EMNLP}, \textsc{GraphQuestions}
\cite{su2016generating}, and \textsc{Spades}
\cite{bisk2016evaluating}.  Experimental results across datasets show
that our weakly-supervised semantic parser achieves state-of-the-art
performance.

\section{The Neural Parser-Ranker}
\label{npr}

Conventional weakly-supervised semantic parsers \cite{liang2016learning} consist of two major
components: a parser, which is
chart-based and non-parameterized, recursively builds derivations for
each utterance span using dynamic programming. A learner, which is a
log-linear model, defines features useful for scoring and ranking the
set of candidate derivations, based on the correctness of
execution results.  As mentioned in
\newcite{liang2016learning}, the chart-based parser brings a
disadvantage since it does not support incremental contextual
interpretation. The dynamic programming algorithm requires that
features of a span are defined over sub-derivations in that span.

In contrast to a chart-based parser, a parameterized neural semantic
parser decodes logical forms with global utterance features.
However, training a weakly-supervised neural parser is challenging since there is no access
to gold-standard logical forms for backpropagation. 
Besides,
it should be noted that a neural decoder is conditionally
generative: decoding is performed greedily conditioned on the utterance and the
generation history---it makes no use of global
logical form features. In this section,
we introduce a parser-ranker framework which combines the best of
conventional and neural approaches in the context of weakly-supervised
semantic parsing.

\subsection{Parser \label{parser}}

Our work follows \newcite{cheng2017learning,cheng2017learning2} in using LISP-style
functional queries as the logical formulation. Advantageously,
functional queries are recursive, tree-structured and can naturally
encode logical form derivations (i.e.,~functions and their application
order).  For example, the utterance ``\textsl{who is obama's eldest
  daughter}'' is simply represented with the function-argument
structure \lform{argmax(daughterOf(Obama), ageOf)}.  Table
\ref{funql} displays the functions we use in this work; a more detailed specifications can be found in the appendix.

To generate logical forms, our system adopts a variant of the neural
sequence-to-tree model proposed in \newcite{cheng2017learning}.
During generation, the prediction space is restricted by the grammar
of the logical language (e.g.,~the type and the number of arguments
required by a function) in order to ensure that output logical forms
are well-formed and executable.  The parser consists of a
bidirectional LSTM \cite{hochreiter1997long} encoder and a stack-LSTM
\cite{dyer2015transition} decoder, introduced as follows.

\begin {table*}[t]
\begin{center}
	\footnotesize
	\begin{tabular}{|p{1.3cm} | p{7cm} | p{5cm}|}
		\hline
        \multicolumn{1}{|c|}{Function} & \multicolumn{1}{c|}{Utility} & \multicolumn{1}{c|}{Example} \\ \hline
		\lform{findAll} &  returns the entity set of a given type  & \textsl{find all mountains} \newline \lform{findAll(mountain)} \\ \hline
		\lform{filter$_=$} \newline \lform{filter$_<$}
          \newline \lform{filter$_>$} & \vspace*{.145cm} filters an entity set
                                        with constraints  & \textsl{all mountains in Europe} \newline \lform{filter$_=$(findAll(mountain), mountain\_location, Europe)}\\ \hline
		\lform{count} & computes the cardinality of an entity set & \textsl{how many mountains are there} \newline \lform{count(findAll(mountain))} \\  \hline
		\lform{argmax} \newline \lform{argmin}  & finds  the subset of an entity set whose certain property is maximum (or minimum)& \textsl{the highest mountain} \newline \lform{argmax(findAll(mountain), mountain\_altitude)} \\ \hline
		\lform{relation} & denotes a KB relation; in generation, \lform{relation} acts as  placeholder for all relations &  \textsl{height of mountain} \newline \lform{mountain\_altitude} \\ \hline
		\lform{entity} & denotes a KB entity;  in generation, \lform{entity} acts as placeholder for all entities  & \textsl{Himalaya} \newline \lform{Himalaya}\\
		\hline
	\end{tabular}
\end{center}
\vspace{-2ex}
\caption{List of functions supported by our functional query language,
  their utility, and examples.\label{funql}} 
%\vspace{-2ex}
\end{table*}

\paragraph{Bidirectional-LSTM Encoder} 
\label{sec:encoder}

The  bidirectional LSTM  encodes a
variable-length utterance $x=(x_1, \cdots, x_n)$ into a list of token
representations $[h_1, \cdots, h_n]$, where each representation is the
concatenation of the corresponding forward and backward LSTM states.

\paragraph{Stack-LSTM Decoder}  
After the utterance is encoded, the logical form is generated with a
stack-LSTM decoder.  The output of the decoder consists of functions
which generate the logical form as a derivation tree in depth-first
order. There are three classes of functions:

\begin{itemize}
\item \normalfont{\textit{Class-1}} functions generate non-terminal
  tree nodes. In our formulation, non-terminal nodes include
  language-dependent functions such as \lform{count} and
  \lform{argmax}, as described in the first four rows of
  Table~\ref{funql}.  A special non-terminal node is the relation
  placeholder \lform{relation}.

\item \normalfont{\textit{Class-2}} functions generate terminal tree
  nodes. In our formulation, terminal nodes include the relation
  placeholder \lform{relation} and the entity placeholder
  \lform{entity}.

\item \normalfont{\textit{Class-3}} function \lform{reduce}
  completes a subtree. Since generation is performed in depth-first
  order, the parser needs to identify when the generation of a subtree
  completes, i.e.,~when a function has seen all its required
  arguments.
\end{itemize} 

\begin{figure}[t]
\begin{tabular}{p{7cm}} \hline
\begin{minipage}[c]{.5\linewidth}
		\begin{tikzpicture}[scale=1]
		\tikzset{every tree node/.style={align=center,anchor=base}}
		\tikzset{level 1+/.style={level distance=2.5\baselineskip}}
		\Tree [ .\lform{argmax} [.\lform{relation-daughterOf}
		[.\lform{entity-Barack\_Obama} ]
		 ] \lform{relation-ageOf} ]
		\end{tikzpicture} 
\end{minipage}\\\hline
\small{Functions for generation (parser)}:  \lform{\small argmax,
  relation, entity, reduce, relation, reduce} \\
	\small{Functions for encoding (inverse parser)}:
  \lform{\small entity, relation, reduce, relation, argmax, reduce}\\ \hline
\end{tabular}
\caption{Derivation tree for the utterance ``\textsl{who is obama's
    eldest daughter}'' (top), and corresponding functions for
  generation and encoding (bottom).  \label{tree}}
\end{figure}
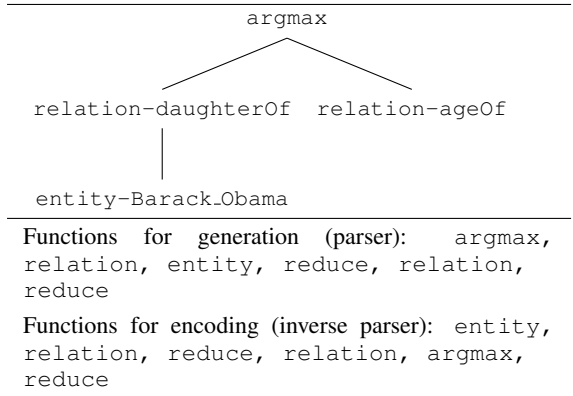

The functions used to generate the example logical form
\lform{argmax(daughterOf(Obama), ageOf)} are shown in
Figure~\ref{tree}.  The stack-LSTM makes two types of updates based on
the functions it predicts:

\begin{itemize} 
\item \normalfont{\textit{Update-1}:} when a \textit{Class-1} or
  \textit{Class-2} function is called, a non-terminal or terminal
  token $l_t$ will be generated, At this point, the stack-LSTM state,
  denoted by~$g_t$, is updated from its older state $g_{t-1}$ as in an
  ordinary LSTM:
\begin{equation}
g_t = \textnormal{LSTM} (l_t, g_{t-1})
\end{equation}
The new state is additionally pushed onto the stack marking whether it
corresponds to a non-terminal or terminal.  

\item \normalfont{\textit{Update-2}}: when the \lform{reduce}
  function is called (\textit{Class-3}), the states of the stack-LSTM
  are recursively popped from the stack until a non-terminal is
  encountered. This non-terminal state is popped as well, after which
  the stack-LSTM reaches an intermediate state denoted
  by~$g_{t-1:t}$. At this point, we compute the representation of the
  completed subtree~$z_t$ as:
\begin{equation}
z_t = W_z \cdot[p_z : c_z]
\end{equation}
where $p_z$ denotes the parent (non-terminal) embedding of the
subtree, and $c_z$ denotes the average embedding of the children
(terminals or already-completed subtrees). $W_z $ is the weight
matrix.  Finally, $z_t$~serves as input for updating~$g_{t-1:t}$
to~$g_t$:
\begin{equation}
g_t = \textnormal{LSTM} (z_t, g_{t-1:t})
\label{reduce}
\end{equation}

\end{itemize} 

\paragraph{Prediction}  
At each time step of the decoding, the parser first predicts a
subsequent function $f_{t+1}$ conditioned on the decoder state~$g_t$
and the encoder states~$h_1 \cdots h_n$.  We apply standard soft
attention \cite{bahdanau2014neural} between~$g_t$ and the encoder
states~$h_1 \cdots h_n$ to compute a feature representation
$\bar{h}_t $:
\begin{equation}
u_t^i  = V \tanh (W_h h_i + W_g g_t) 
\end{equation}
\begin{equation}
a_t^i  =  \textnormal{softmax} (u_t^i )
\end{equation}
\begin{equation}
\bar{h}_t  = \sum_{i=1}^n  a_t^i  h_i 
\label{attention}
\end{equation}
where $V$, $W_h$, and $W_g$ are all weight parameters.  The prediction
of the function~$f_{t+1}$ is computed with a softmax classifier, which
takes the concatenated features $\bar{h}_t $ and $g_t$ as input:
\begin{equation}
f_{t+1}  \sim \textnormal{softmax} (  W_{y} \tanh( W_f [\bar{h}_t, g_t] )  )
\label{p1}
\end{equation}
where $W_y$ and $W_f$ are weight parameters. When~$f_{t+1}$ is a
language-dependent function (first four rows in Table~\ref{funql},
e.g., \texttt{argmax}), it is directly used as a non-terminal token
$l_{t+1}$ to construct the logical form.  However, when~$f_{t+1}$ is a
\lform{relation} or \lform{entity} placeholder, we
further predict the specific relation or entity~$l_{t+1}$ with another
set of neural parameters:
\begin{equation}
l_{t+1}  \sim \textnormal{softmax} (  W_{y'} \tanh( W_{l} [\bar{h}_t, g_t] )  )
\label{p2}
\end{equation}
where $W_{y'}$ and $W_{l'}$ are weight matrices.

Note that in the weakly supervised setting, the parser decodes a list
of candidate logical forms~$Y$ with beam search, instead of outputting
the most likely logical form~$y$.  During training, candidate logical
forms are executed against a knowledge base to find those which are
consistent (denoted by $Y_c(x)$) and lead to the correct
denotation. Then, the parser is trained to maximize the total log
likelihood of these consistent logical forms:
\begin{equation}
\begin{split}
& \sum_{y \in Y_c(x)} \log p(y|x)  =  \\
& \sum_{y \in Y_c(x)} \log p(f_1,\cdots, f_k, l_1, \cdots, l_o|x) 
\end{split}
\label{obj_parser}
\end{equation}
where $k$ denotes the number of functions used to generate the logical form, and $o$ (smaller than $k$) denotes the number of tree nodes in the logical form.

\subsection{Ranker}
\label{sec:ranker}

It is impractical to rely solely on a neural decoder to find the most
likely logical form at run time in the weakly-supervised setting.  One
reason is that although the decoder utilizes global utterance features
for generation, it cannot leverage global features of the
logical form since a logical form is conditionally generated following a
specific tree-traversal order. To this end, we follow previous work
\cite{berant2013semantic} and introduce a ranker to the system.  The
role of the ranker is to score the candidate logical forms generated
by the parser; at test time, the logical form receiving the highest
score will be used for execution.  The ranker is a discriminative
log-linear model over logical form~$y$ given utterance~$x$:
\begin{equation}
\log_\theta p(y|x) = \frac{\exp (\phi(x, y)^T \theta)}{\sum_{y' \in Y(x)} \exp(\phi(x, y')^T \theta)}
\end{equation}
where $Y(x)$ is the set of candidate logical forms; $\phi$ is the
feature function that maps an utterance-logical form pair onto a
feature vector; and $\theta$ denotes the weight parameters of the
model.

Since the training data consists only of utterance-denotation pairs,
the ranker is trained to maximize the log-likelihood of the correct
answer~$z$ by treating logical forms as a latent variable:
\begin{equation}
\log p(z|x) = \log \sum_{y \in Y_c(x)} p(y|x) p(z|x,y)
\label{obj_ranker}
\end{equation}
where $Y_c(x)$ denotes the subset of candidate logical forms which
execute to the correct answer; and $p(z|x,y)$ equates to 1 in this
case.

Training of the neural parser-ranker system involves the following
steps. Given an input utterance, the parser first generates a list of
candidate logical forms via beam search.  The logical forms are then
executed and those which yield the correct denotation are marked as
consistent.  The parser is trained to optimize the total likelihood of
consistent logical forms (Equation~\eqref{obj_parser}), while the
ranker is trained to optimize the marginal likelihood of denotations
(Equation~\eqref{obj_ranker}).  The search space can be further
reduced by performing entity linking which restricts the number of
logical forms to those containing only a small set of entities.

\section{Handling Spurious Logical Forms}
The neural parser-ranker system relies on beam search to find
consistent logical forms that execute to the correct answer. These
logical forms are then used as surrogate annotations and provide
supervision to update the parser's parameters. However, some of these
logical forms will be misleading training signals for the neural
semantic parser on account of being spurious: they coincidentally
execute to the correct answer without matching the utterance
semantics.

In this section we propose a method of removing spurious logical forms
by validating how well they match the utterance meaning. The intuition
is that a meaning-preserving logical form should be able to
reconstruct the original utterance with high likelihood.  However,
since spurious logical forms are not annotated either, a direct
maximum likelihood solution does not exist.  To this end, we propose a
generative model for measuring the \emph{reconstruction} likelihood.

The model assumes utterance~$x$ is generated from the
corresponding logical form~$y$, and only the utterance is observable.
The objective is therefore to maximize the log marginal likelihood
of~$x$:
\begin{equation}
\log p(x) = \log \sum_y p(x, y)
\end{equation}
We adopt neural variational inference \cite{mnih2014neural} to solve the above objective, which is equivalent to maximizing an
evidence lower bound:
\begin{equation}
\vspace*{-.8cm}
\end{equation}
\[\begin{split}
\log p(x) & = \log  \frac{q(y|x) p(x|y) p(y)}{q(y|x)} \\ 
& \geq \mathbb{E}_{q(y|x)} \log p(x|y) + \mathbb{E}_{q(y|x)} \log \frac{p(y)}{q(y|x)} \\
 \end{split}
\]

Since our semantic parser always outputs well-formed logical forms, we assume
a uniform constant prior~$p(y)$.  The
above objective can be thus reduced to:
\begin{equation}
\hspace*{-.33cm}\mathbb{E}_{q(y|x)} \log p(x|y) - \mathbb{E}_{q(y|x)} \log q(y|x) = \mathcal{L}(x)
\label{obj1}
\end{equation}
where the first term computes the reconstruction likelihood $p(x|y)$;
and the second term is the entropy of the approximated posterior
$q(y|x) $ for regularization.  Specifically, we use the semantic
parser to compute the approximated posterior $q(y|x)$.\footnote{In
  Section~\ref{parser}, we used a different notation for the
  output distribution of the semantic parser as~$p(y|x)$.}  The
reconstruction likelihood $p(x|y)$ is computed with an inverse parser
which recovers utterance~$x$ from its logical form~$y$.  We
use~$p(x|y)$ to measure how well the logical form reflects the
utterance meaning; details of the inverse parser are described as
follows.

\paragraph{Stack-LSTM Encoder}  
To reconstruct utterance~$x$, logical form~$y$ is first encoded with a
stack-LSTM encoder.  To do that, we deterministically convert the
logical form into a sequence of \textit{Class-1} to \textit{Class-3}
functions, which correspond to the creation of tree nodes and
subtrees.  Slightly different from the top-down generation process, the
functions here are obtained in a bottom-up order to facilitate
encoding.  Functions used to encode the example logical form
\lform{argmax(daughterOf(Obama), ageOf)} are shown in
Figure~\ref{tree}.

The stack-LSTM sequentially processes the functions and updates its
states based on the class of each function, following the same
principle (\textit{Update-1} and \textit{Update-2}) described in
Section~\ref{parser}.  We save a list of terminal, non-terminal and
subtree representations $[g_1, \cdots, g_s]$, where each
representation is the stack-LSTM state at the corresponding time step
of encoding.  The list essentially contains the representation of
every tree node and the representation of every subtree (the total
number of representations is denoted by $s$).

\paragraph{LSTM Decoder}  
Utterance~$x$ is reconstructed with a standard LSTM decoder attending
to tree nodes and subtree representations.  At each time step, the
decoder applies  attention between decoder state~$r_t$ and tree fragment
representations $[g_1, \cdots, g_s]$:
\begin{equation}
v_t^i  = V' \tanh (W_{g'} g_i + W_r r_t) 
\end{equation}
\begin{equation}
b_t^i  =  \textnormal{softmax} (v_t^i )
\end{equation}
\begin{equation}
\bar{g}_t  = \sum_{i=1}^s  b_t^i  g_i 
\label{attention2}
\end{equation}
and predicts the probability of the next word as:
\begin{equation}
x'_{t+1}  \sim \textnormal{softmax} (  W_{x'} \tanh( W_{f'} [\bar{g}_t, r_t] )  )
\label{p22}
\end{equation}
where $W$s and $V'$ are all weight parameters. 

\paragraph{Gradients}
The training objective of the generative model is given in
Equation~\eqref{obj1}. The parameters of the neural network include
those of the original semantic parser (denoted by $\theta$) and the
inverse parser (denoted by $\phi$).  The gradient of
Equation~\eqref{obj1} with respect to $\phi$ is:
\begin{equation}
	\frac{\partial \mathcal{L}(x)}{\partial \phi} = \mathbb{E}_{q(y|x)} \frac{\partial \log p(x|y)}{\partial \phi}
\end{equation}
and the gradient with respect to $\theta$ is:
\vspace{1cm}
\begin{equation}
\vspace*{-2cm}
\end{equation}
\[
\begin{split}
&\frac{\partial \mathcal{L}(x)}{\partial \theta} =  \mathbb{E}_{q(y|x)}[(\log p(x|y) - \log q(y|x))\\ 
& \hspace*{3cm}\times \frac{\partial \log q(y|x)}{\partial \theta}]
\end{split}\]

Both gradients involve expectations which we estimate with Monte Carlo
method, by sampling logical forms from the distribution~$q(y|x)$.
Recall that in the parser-ranker framework these samples are obtained
via beam search.

\section{Scheduled Training}
Together with the inverse parser for removing spurious logical forms,
the proposed system consists of three components: a parser which
generates logical forms from an utterance, a ranker which measures the
likelihood of a logical form executing to the \textit{correct
  denotation}, and an inverse parser which measures the degree to
which logical forms are \textit{meaning-preserving} using
reconstruction likelihood.  Our semantic parser is trained following a
scheduled training procedure, balancing the two objectives.

\begin{itemize}
\item{\emph{Phase 1}}: at the beginning of training when all model
  parameters are far from optimal, we train only the parser and the
  ranker as described in Section~\ref{npr}; the parser generates a
  list of candidate logical forms, we find those which are consistent
  and update both the parser and the ranker.

\item{\emph{Phase 2}}: we turn on the inverse parser and update
  all three components in one epoch.  However, the reconstruction loss
  is only used to update the inverse parser and we prevent it from
  back-propagating to the semantic parser. This is because at this
  stage of training the parameters of the inverse parser are
  sub-optimal and we cannot obtain an accurate approximation of the
  reconstruction loss.

\item{\emph{Phase 3}}: finally, we allow the reconstruction loss to
  back-propagate to the parser, and all three components are updated
  as normal.  Both training objectives are enabled, the system
  maximizes the likelihood of consistent logical forms
  and the reconstruction likelihood.
\end{itemize}

\section{Neural Lexicon Encoding}
In this section we further discuss how the semantic parser presented
so far can be enhanced with a lexicon. A lexicon is essentially a coarse mapping between
natural language phrases and knowledge base relations and entities,
and has been widely used in conventional chart-based parsers
\cite{berant-EtAl:2013:EMNLP,reddy2014large}. Here, we show how a
lexicon (either hard-coded or statistically-learned \cite{krishnamurthy2016probabilistic}) can be used to benefit a neural semantic parser.

The central idea is that relations or
entities can be viewed as a single-node tree-structured logical form.
For example, based on the lexicon, the natural language phrase
``\textsl{is influenced by}'' can be parsed to the logical form
\lform{influence.influence\_node.influenced\_by}.  We can therefore
pretrain the semantic parser (and the inverse parser) with these basic
utterance-logical form pairs which act as important prior knowledge
for initializing the distributions $q(y|x)$ and $p(x|y)$.  With
pre-trained word embeddings capturing linguistic regularities on the
natural language side, we also expect the approach to help the neural
model generalize to unseen natural language phrases quickly.  For
example, by encoding the mapping between the natural language phrase
``\textsl{locate in}'' and the Freebase predicate
\lform{fb:location.location.containedby}, the parser can potentially
link the new phrase ``\textsl{located at}'' to the same predicate.  We
experimentally assess whether the neural lexicon enhances the
performance of our semantic parser.

\section{Experiments}
\label{sec:experiments}

In this section we evaluate the performance our semantic parser.  We
introduce the various datasets used in our experiments, training
settings, model variants used for comparison, and finally present and
analyze our results.

\subsection{Datasets}
We evaluated our model on three Freebase datasets:
\textsc{WebQuestions} \cite{berant-EtAl:2013:EMNLP},
\textsc{GraphQuestions} \cite{su2016generating} and \textsc{Spades}
\cite{bisk2016evaluating}.  \textsc{WebQuestions} contains~5,810 real
questions asked by people on the web paired by answers.
\textsc{GraphQuestions} contains 5,166 question-answer pairs which
were created by showing 500~Freebase graph queries to Amazon
Mechanical Turk workers and asking them to paraphrase them into
natural language.  \textsc{Spades} contains~93,319 question-answer
pairs which were created by randomly replacing entities in declarative
sentences with a blank symbol.

\subsection{Training}
Across training regimes, the dimensions of word vector, logical form
token vector, and LSTM hidden states (for the semantic parser and the
inverse parser) are 50, 50, and~150, respectively.  Word embeddings
were initialized with Glove embeddings \cite{pennington2014glove}.
All other embeddings were randomly initialized.  We used one LSTM
layer in the forward and backward directions.  Dropout was used before
the softmax activation (Equations \eqref{p1}, \eqref{p2}, and
\eqref{p22}).  The dropout rate was set to~0.5.  Momentum SGD
\cite{sutskever2013importance} was used as the optimization method to
update the parameters of the model.

As mentioned earlier, we use entity linking to reduce the beam search
space.  Entity mentions in \textsc{Spades} are automatically annotated
with Freebase entities \cite{gabrilovich2013facc1}. For
\textsc{WebQuestions} and \textsc{GraphQuestions} we perform entity
linking following the procedure described in
\citet{reddy2016transforming}. We identify potential entity spans
using seven handcrafted part-of-speech patterns and associate them
with Freebase entities obtained from the Freebase/KG
API.\footnote{\url{http://developers.google.com/freebase/}} We use a
structured perceptron trained on the entities found in
\textsc{WebQuestions} and \textsc{GraphQuestions} to select the top~10
non-overlapping entity disambiguation possibilities.  We treat each
possibility as a candidate entity and construct candidate utterances
with a beam search of size~300.

Key features of the log-linear ranker introduced in Section~\ref{npr}
include the entity score returned by the entity linking system, the
likelihood score of the relation in the logical form predicted by the
parser, the likelihood score of the the logical form predicted by the
parser, the embedding similarity between the relation in the logical
form and the utterance, the similarity between the relation and the
question words in the utterance, and the answer type as indicated by
the last word in the Freebase relation \cite{xu2016question}.  All
features are normalized across candidate logical forms.  For all
datasets we use average~F1 \cite{berant-EtAl:2013:EMNLP} as our
evaluation metric.

\subsection{Model Variants}
\label{sec:model-variants}

We experiment with three variants of our model.  We primarily consider
the neural parser-ranker system (denoted by \textsc{npr}) described in
Section~\ref{npr} which is trained to maximize the likelihood of
consistent logical forms.  We then compare it to a system augmented
with a generative ranker (denoted by \textsc{granker}), introducing
the second objective of maximizing the reconstruction likelihood.
Finally, we examine the impact of neural lexicon encoding when it is
used for the generative ranker, and also when it is used for the
entire system.

\subsection{Results}
\label{sec:experimental-results}

\begin {table}[t]
\begin{center}
	\begin{tabular}{|lc|}
		\hline
		Models & F1 \\ \hline
		\newcite{berant-EtAl:2013:EMNLP} & 35.7\\
		
		\newcite{berant2014semantic} & 39.9 \\
		
		\newcite{berant2015imitation} & 49.7 \\
		\newcite{reddy2016transforming} & 50.3 \\ \hline \hline
		
		\newcite{yao2014information} & 33.0 \\
		\newcite{bast2015more} & 49.4 \\ \hline \hline
		\newcite{bordesquestion} & 39.2 \\
		\newcite{dong2015question} & 40.8 \\
		\newcite{yih2015semantic} & 52.5 \\
		\newcite{xu2016question} & 53.3 \\
		\newcite{cheng2017learning} & 49.4 \\
		\textsc{npr} & 50.1  \\
		 \, \, + \small{\textsc{granker} }& 50.2 \\
		 \, \, +  \small{lexicon encoding on \textsc{granker}}  & 51.7 \\
		 \, \, +  \small{lexicon encoding on parser and \textsc{granker}}  & 52.5 \\
		\hline
	\end{tabular}
\end{center}
\caption{\textsc{WebQuestions} results.}
\label{webqa}
\end{table}

Experimental results on \textsc{WebQuestions} are shown in
Table~\ref{webqa}.  We compare the performance of \textsc{npr} with
previous work, including conventional chart-based semantic parsing
models (e.g., \newcite{berant-EtAl:2013:EMNLP}; first block in
Table~\ref{webqa}), information extraction models (e.g.,
\newcite{yao2014information}; second block in Table~\ref{webqa}), and
more recent neural question-answering models (e.g.,
\newcite{dong2015question}; third block in Table~\ref{webqa}).  Most
neural models do not generate logical forms but instead build a
differentiable network to solve a specific task such as
question-answering.  An exception is the neural sequence-to-tree model
of \newcite{cheng2017learning}, which we extend to build the vanilla
\textsc{npr} model. A key difference of \textsc{npr} is that it
employs soft attention instead of hard attention, which is
\newcite{cheng2017learning} use to rationalize predictions.

As shown in Table~\ref{webqa}, the basic \textsc{npr} system
outperforms most previous chart-based semantic parsers. Our results
suggest that neural networks are powerful tools for generating
candidate logical forms in a weakly-supervised setting, due to their
ability of encoding and utilizing sentential context and generation
history.  Compared to \newcite{cheng2017learning}, our system also
performs better. We believe the reason is that it employs soft
attention instead of hard attention. Soft attention makes the parser
fully differentiable and optimization easier.  The addition of the
inverse parser ($+$\textsc{granker}) to the basic \textsc{npr} model
yields marginal gains while the addition of the neural lexicon
encoding to the inverse parser brings performance improvements over
\textsc{npr} and \textsc{granker}. We hypothesize that this is because
the inverse parser adopts an unsupervised training objective, which
benefits substantially from prior domain-specific knowledge used to
initialize its parameters.  When neural lexicon encoding is
incorporated in the semantic parser as well, system performance can be
further improved.  In fact, our final system (last row in
Table~\ref{webqa}) outperforms all previous models except that of
\newcite{xu2016question}, which uses external Wikipedia resources to
prune out erroneous candidate answers.

\begin {table}[t!]
\begin{center}
	\begin{tabular}{|lr|}
		\hline
		{Models} & \multicolumn{1}{c|}{F1} \\ \hline 
		\textsc{sempre} \cite{berant-EtAl:2013:EMNLP} & 10.80 \\ %
		\textsc{parasempre} \cite{berant2014semantic} & 12.79 \\
		\textsc{jacana} \cite{yao2014information} & 5.08 \\
		\textsc{ScanneR} \cite{cheng2017learning} & 17.02 \\
		\textsc{udeplambda} \cite{reddy2017universal} & 17.70 \\
		\textsc{npr}  & 17.30  \\
		\, \, + \small{\textsc{granker}} & 17.33\\
		\, \, +  \small{lexicon encoding on \textsc{granker}}  & 17.67 \\
		\, \, +  \small{lexicon encoding on parser and \textsc{granker}}  & 18.22 \\
		\hline
	\end{tabular}
\end{center}
\caption{\textsc{GraphQuestions} results.} 
\label{gqa}
\vspace{-2ex}
\end{table}

\begin {table}[t]
\begin{center}
	\begin{tabular}{|lc|}
		\hline 
		{Models} & F1 \\ \hline
		Unsupervised CCG \cite{bisk2016evaluating} & 24.8 \\
		Semi-supervised CCG \cite{bisk2016evaluating} & 28.4 \\
		Supervised CCG \cite{bisk2016evaluating} & 30.9 \\
		Rule-based system \cite{bisk2016evaluating} & 31.4 \\
		Sequence-to-tree \cite{cheng2017learning} & 31.5 \\
		Memory networks \cite{das2017question} & 39.9 \\
		\textsc{npr}  & 32.4  \\
		\, \, + \small{\textsc{granker}} & 33.1 \\
		\, \, +  \small{lexicon encoding on \textsc{granker}}  & 35.5\\
		\, \, +  \small{lexicon encoding on parser and \textsc{granker} } & 37.6 \\
		\hline
	\end{tabular}
\end{center}
\caption{\textsc{Spades} results.}
\label{spade}
\end{table}

Tables~\ref{gqa} and~\ref{spade} present our results on
\textsc{GraphQuestions} and \textsc{Spades}, respectively. Comparison
systems for \textsc{GraphQuestions} include two chart-based semantic
parsers \cite{berant-EtAl:2013:EMNLP,berant2014semantic}, an
information extraction model \cite{yao2014information}, a neural
sequence-to-tree model with hard attention \cite{cheng2017learning}
and a model based on universal dependency to logical form conversion
\cite{reddy2017universal}.  On \textsc{Spades} we compare with the
method of \newcite{bisk2016evaluating} which parses an utterance into
a syntactic representation which is subsequently grounded to Freebase;
and also with \newcite{das2017question} who employ memory networks and
external text resources.  Results on both datasets follow similar
trends as in \textsc{WebQuestions}. The best performing \textsc{npr}
variant achieves state-of-the-art results on \textsc{GraphQuestions}
and it comes close to the best model on \textsc{Spades} without using
any external resources.

\begin {table*}[t]
\begin{center}
	\begin{tabular}{|@{}l@{}|}
		\hline 
\emph{which baseball teams were coached by dave eiland}\\
         \lform{ 
           \textcolor{red}{baseball.batting\_statistics.player:baseball.batting\_statistics.team(ent.m.0c0x6v)}} \\
         \lform{ 
           \textcolor{blue}{baseball.historical\_coaching\_tenure.baseball\_coach:baseball.historical\_coaching\_tenure.}}\\
\lform{ 
           \textcolor{blue}{baseball\_team(ent.m.0c0x6v)}}\\ \hline
{\emph{who are coca-cola's endorsers}}\\
         \textcolor{red}{\lform{ 
             food.nutrition\_fact.food:food.nutrition\_fact.nutrient(ent.m.01yvs)}} \\
         \textcolor{blue}{\lform{
             business.product\_endorsement.product:business..product\_endorsement.endorser(ent.m.01yvs)}}\\\hline
         \emph{what are the aircraft models that are
           comparable 
          to airbus 380} \\ 
         \textcolor{red}{\lform{aviation.aviation\_incident\_aircraft\_relationship.flight\_destination:aviation.aviation\_}}\\
\textcolor{red}{\lform{incident\_aircraft\_relationship.aircraft\_model(ent.m.0qn2v)}} \\
         \textcolor{blue}{\lform{aviation.comparable\_aircraft\_relationship(ent.m.018rl2)}} \\         
        \hline
	\end{tabular}
\end{center}
\caption{Comparison between logical forms preferred by \textsc{npr}
  before and after the addition of the inverse parser. Spurious
  logical forms (red color) receive higher scores than
  semantically-correct ones (blue color). The scores of these spurious
  logical forms decrease when they are explicitly handled.}
\label{qualitive}
\end{table*}

One of the claims put forward in this paper is that the extended
\textsc{npr} model reduces the impact of spurious logical forms during
training.  Table~\ref{qualitive} highlights examples of spurious
logical forms which are not semantically correct but are nevertheless
assigned higher scores in the vanilla \textsc{npr} (red colour). These
logical forms become less likely in the extended \textsc{npr}, while
the scores of more semantically faithful representations (blue
colour) are boosted.

%compared to more semantically-correct
%counterparts. These spurious logical forms are assigned higher scores
%in the vanilla \textsc{npr}, but lower scores in the extended
%\textsc{npr}.

\subsection{Discussion} 

The vanilla \textsc{npr} model is optimized with consistent logical
forms which lead to correct denotations.  Although it achieves
competitive results compared to chart-based parsers, the training of
this model can be misled by spurious logical forms.  The introduction
of the inverse parser aims to alleviate the problem by scoring how a
logical form reflects the utterance semantics.  Although the inverse
parser is not directly used to rank logical forms at test time, the
training objective it adopts encourages the parser to generate
meaning-preserving logical forms with higher likelihood. These
probabilities are used as features in the log-linear ranker, and
therefore the inverse parser affects the ranking results, albeit
implicitly.  

However, we should point out that the unsupervised training objective
is relatively difficult to optimize, since there are no constraints to
regularize the latent logical forms. This motivates us to develop a
scheduled training procedure; as our results show, when trained
properly the inverse parser and the unsupervised objective bring
performance gains. Moreover, the neural lexicon encoding method we
applied essentially produces synthetic data to further regularize the
latent space.  %With pretrained word embeddings capture phrase
%similarities, the method also helps the parser to generalize to unseen
%phrases quickly. 

\section{Related Work}
Various types of supervision have been explored to train semantic
parsers. Early semantic parsers have used annotated training data
consisting of sentences and their corresponding logical forms
\cite{kate2006using,kate2005learning,lu2008generative,kwiatkowksi2010inducing}.
In order to scale semantic parsing to open-domain problems,
weakly-supervised semantic parsers are trained on utterance-denotation
pairs
\cite{liang2011learning,krishnamurthy2012weakly,berant2013semantic,choi_scalable_2015,krishnamurthy2015learning,pasupat2016inferring,gardner_openvocabulary_2017,reddy2017universal}.
Most previous work employs a chart-based parser to produce logical
forms from a grammar which combines domain-general aspects with
lexicons.

Recently, neural semantic parsing has attracted a great deal of
attention.  Previous work has mostly adopted fully-supervised,
sequence-to-sequence models to generate logical form strings from
natural language utterances
\cite{dong2016language,jia2016data,kovcisky2016semantic}.  Other work
explores the use of reinforcement learning to train neural semantic
parsers from question-answer pairs \cite{liang2016neural} or from user
feedback \cite{iyer2017learning}. More closely related to our work,
\newcite{goldman2018weakly} adopt a neural semantic parser and a
discriminative ranker to solve a visual reasoning challenge. They
attempt to alleviate the search space and spuriousness challenges with
abstractive examples.  \newcite{yin2018structvae} adopt a tree-based
variational autoencoder for semi-supervised semantic parsing.  Neural
variational inference has also been used in other NLP tasks including
relation discovery \cite{marcheggiani2016discrete}, sentence compression
\cite{miao2016language}, and parsing \cite{cheng2017generative}.

\section{Conclusions}
In this work we proposed a weakly-supervised neural semantic parsing
system trained on utterance-denotation pairs.  The system employs a
neural sequence-to-tree parser to generate logical forms for a natural
language utterance.  The logical forms are subsequently ranked with
two components and objectives: a log-linear model which scores the
likelihood of correct execution, and a generative neural inverse
parser which measures whether logical forms are meaning preserving. We
proposed a scheduled training procedure to balance the two objectives,
and a neural lexicon encoding method to initialize model parameters
with prior knowledge. Experiments on three semantic parsing datasets
demonstrate the effectiveness of our system. In the future, we would
like to train our parser with other forms of supervision such as
feedback from users \cite{he2016human,iyer2017learning} or
textual evidence \cite{yin2018structvae}.

\paragraph{Acknowledgments} 
This research is supported by a Google PhD Fellowship and an AdeptMind
Scolar Fellowship to the first author.  We also gratefully acknowledge
the financial support of the European Research Council (award number
681760; Lapata).

\bibliographystyle{acl_natbib_nourl} 
\bibliography{conll2018}
\end{document}